\documentclass[twocolumn, switch]{article} 

\usepackage{preprint}

\usepackage{amsmath, amsthm, amssymb, amsfonts}

\usepackage[numbers,square]{natbib}
\usepackage[utf8]{inputenc}	
\usepackage[T1]{fontenc}	
\usepackage{xcolor}		
\usepackage[colorlinks = true,
            linkcolor = purple,
            urlcolor  = blue,
            citecolor = cyan,
            anchorcolor = black]{hyperref}	
\usepackage{booktabs} 		
\usepackage{nicefrac}		
\usepackage{microtype}		
\usepackage{lineno}		
\usepackage{float}			

\usepackage{lipsum}		

\usepackage{newfloat}
\DeclareFloatingEnvironment[name={Supplementary Figure}]{suppfigure}
\usepackage{sidecap}
\sidecaptionvpos{figure}{c}

\graphicspath{ {./images/} }

\usepackage{titlesec}
\titlespacing\section{0pt}{12pt plus 3pt minus 3pt}{1pt plus 1pt minus 1pt}
\titlespacing\subsection{0pt}{10pt plus 3pt minus 3pt}{1pt plus 1pt minus 1pt}
\titlespacing\subsubsection{0pt}{8pt plus 3pt minus 3pt}{1pt plus 1pt minus 1pt}

\usepackage{tikz,xcolor,hyperref}

\definecolor{lime}{HTML}{A6CE39}
\DeclareRobustCommand{\orcidicon}{
	\begin{tikzpicture}
	\draw[lime, fill=lime] (0,0) 
	circle [radius=0.16] 
	node[white] {{\fontfamily{qag}\selectfont \tiny ID}};
	\draw[white, fill=white] (-0.0625,0.095) 
	circle [radius=0.007];
	\end{tikzpicture}
	\hspace{-2mm}
}
\foreach \x in {A, ..., Z}{\expandafter\xdef\csname orcid\x\endcsname{\noexpand\href{https://orcid.org/\csname orcidauthor\x\endcsname}
			{\noexpand\orcidicon}}
}

\title{Autonomous Driving Implementation in an Experimental Environment}

\usepackage{xwatermark}
\newwatermark[firstpage,color=gray!90,angle=0,scale=0.28, xpos=0in,ypos=-5in]{*correspondence: \texttt{namig.aliyev[at]ogr.sakarya.edu.tr}}

\usepackage{authblk}

\author[1\thanks{\tt{namig.aliyev@ogr.sakarya.edu.tr}}]{Namig Aliyev\orcidA{}}
\author[2\thanks{\tt{oguzhan.sezer1@ogr.sakarya.edu.tr}}]{Oguzhan Sezer\orcidB{}}
\author[3\thanks{\tt{mehmet.guzel@ogr.sakarya.edu.tr}}]{Mehmet Turan Guzel\orcidC{}}

\affil[1,2,3]{Department of Computer Engineering, Sakarya University}

\begin{document}

\twocolumn[ 
  \begin{@twocolumnfalse} 
  
\maketitle

\begin{abstract}
Autonomous systems require identifying the environment and it has a long way to go before putting it safely into practice. In autonomous driving systems, the detection of obstacles and traffic lights are of importance as well as lane tracking. In this study, an autonomous driving system is developed and tested in the experimental environment designed for this purpose. In this system, a model vehicle having a camera is used to trace the lanes and avoid obstacles to experimentally study autonomous driving behavior. Convolutional Neural Network models were trained for Lane tracking. For the vehicle to avoid obstacles, corner detection, optical flow, focus of expansion, time to collision, balance calculation, and decision mechanism were created,
respectively.
\end{abstract}
\vspace{0.35cm}

  \end{@twocolumnfalse} 
] 



\section{Introduction}
The autonomous driving system is one of the most popular
smart autonomous systems recently. Nowadays, it is aimed
to minimize driver-related errors with autonomous driving
systems. Today, we can say that autonomous driving systems
have speed control with radar and distance sensors, lane
tracking, and lane change after cameras on the vehicle.
Autonomous vehicles are one of the most effective use cases
where hardware and software work together. The hardware
enables the vehicle to move and communicate with a range
of cameras, sensors, while the software processes
information and provides control.

Today, many automobile companies are attempting to
produce cars with autonomous driving systems. We can say
Tesla company as the leading company. The cars they
produce have a full automation driving system. The data set
is collected in real-time from approximately 1 million
vehicles. 70,000 GPU’s are trained per hour. It is capable of
semantic segmentation, object recognition, depth estimation.
There are 1000 different estimates per step each time. Some
companies use the LIDAR device to model depth prediction
and 3D perception. Depth prediction is a fundamental task in
perceiving the 3D environment around us \cite{casser2019depth}.

In this study, lane tracking, which is one of the two most
important abilities in autonomous vehicles, and the ability to
avoid obstacles for the robot vehicle to drive freely without
hitting any obstacle are discussed. The main purpose of this
research was to develop lane tracking and obstacle avoidance
capabilities with different methods and solutions for an
autonomous driving system on an experimental vehicle.

The robot vehicle, remote control module, and experimental
map that constitute the hardware part of the project were
prepared. Raspberry Pi module on the vehicle forms the
brain of the vehicle. Raspberry PI communicates with the
remote-control module via wireless network (RF24) and
computer via embedded software. The Raspberry PI module,
which plays the role of the brain of the system in the later
stages of the project, was renewed with the Coral Dev Board
\cite{cass2019taking} device developed by Google for artificial intelligence
model’s due to its inadequate performance.

Supervised Learning \cite{caruana2006empirical}, \cite{zhu2009introduction} 
approach, which is one of the
Machine Learning \cite{alpaydin2020introduction}, \cite{schalkoff2007pattern}
techniques, was used for lane
tracking. With the help of a remote-control device, the robot
vehicle was moved along the track and dataset collection was
carried out through the camera on the vehicle. This dataset
created consists of images and action information taken at
the time of that image. Then, the images taken were
simplified with image processing techniques, and the strip
lines were brought to the fore. At this stage, Convolutional
Neural Network \cite{albawi2017understanding} was used while creating an artificial
intelligence model. CNN is a type of artificial neural network
developed to solve problems such as image classification,
object detection, and style transfer. Since the images are our
main data source, it was decided to use CNN.

The target problem for avoiding obstacles is the calculation
of contact time or time to collision. The most important
feature focused here was the calculation of the time until the
collision, ie the contact time. In this direction, corner
detection, optical flow focus of expansion, and collision time
were calculated instantaneously on the image taken from the
camera \cite{o2005optical}, \cite{beauchemin1995computation}, \cite{barron2005tutorial}. 
The balance calculation has been made for
the right and left body of the robot vehicle and a decision
mechanism has been created to avoid obstacles. The robot
vehicle has been provided to move without hitting any
obstacle.

\section{Proposed Method}
\label{sec:proposed}

\subsection{Lane Tracking}

In this section, simplification of lane information with
computer vision techniques, and data set collection are given
initially to enable the robot vehicle to move autonomously
by following the lane information on the experimental map.
Next, a detailed description of the designed network
architecture and training is provided. To achieve the
successful model, hyper parameter tuning and data
augmentation, and finally, the testing process is explained.

The dataset collection process will be performed by moving
the robot vehicle over the experimental environment with the
help of a remote control. The images taken from the camera
correctly positioned on the vehicle will first be recorded in
the filing system by simplifying the lane information with
image processing techniques. At the same time, the action
information of the car at the time the image is taken is
recorded in the filing system simultaneously with the images.

\subsubsection{Simplifying Lane Information with Computer
Vision Techniques}

Preparing the images in the data set that we will give to the
neural network in accordance with the purpose is the most
influential factor in the result of the developed neural
network model. If the image is messy, difficult to
understand, and the neural network is not able to distinguish
the features in the image, the error values of the model will
be high and the operation is nothing but a waste of time. For
this purpose, the images taken from the camera on the robot
vehicle were first simplified with image processing
techniques. First, the color space change was performed on
the image. Many color spaces are supported in the OpenCV
library and you can convert between them. In the first step,
the image was converted from RGB color space to grayscale
color space \cite{saravanan2010color}.

\begin{figure}[h]
    \centering
    \includegraphics[scale=0.48]{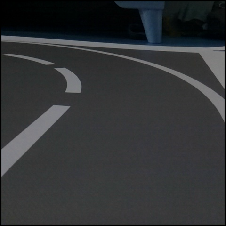}
    \caption{Change from RGB color space to grayscale color
space.}
    \label{fig:fig1}
\end{figure}

Figure 1. shows the image obtained by converting the camera
image taken on the robot vehicle from RGB color space to
grayscale color space. In the image in the grayscale color
space obtained, the stripe lines are desired to be prominent.
With the help of the Canny \cite{xu2017canny} edge detection algorithm,
the strip lines required on the image were made more
prominent.

\begin{figure}[h]
    \centering
    \includegraphics[scale=0.225]{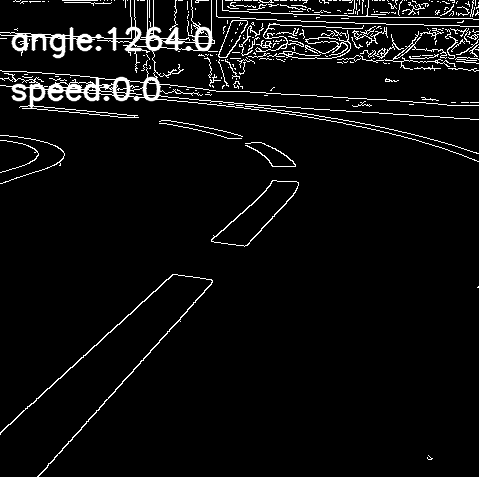}
    \caption{Transformation of grayscale image with Canny edge
detection algorithm.}
    \label{fig:fig2}
\end{figure}

The image taken from the camera has been successfully
simplified and made ready for the use of the neural network.
If you pay attention to the upper left corner of the figure, you
can see the angle and speed, which are the action information
of the vehicle at the time the image is taken from the camera.

\subsubsection{Data Set Collection and Editing}

The collection of the data set will be carried out by moving
the robot vehicle over the experimental environment with the
remote control. In the Supervised Learning machine learning
approach we will use, the data to be trained should be given
to the learner as x and y outputs. While the vehicle is
controlled remotely on the experimental environment we
have prepared before, the image from the camera, the current
servo angle and engine speed are recorded in the filing
system simultaneously.

\begin{figure}[h]
    \centering
    \includegraphics[scale=0.45]{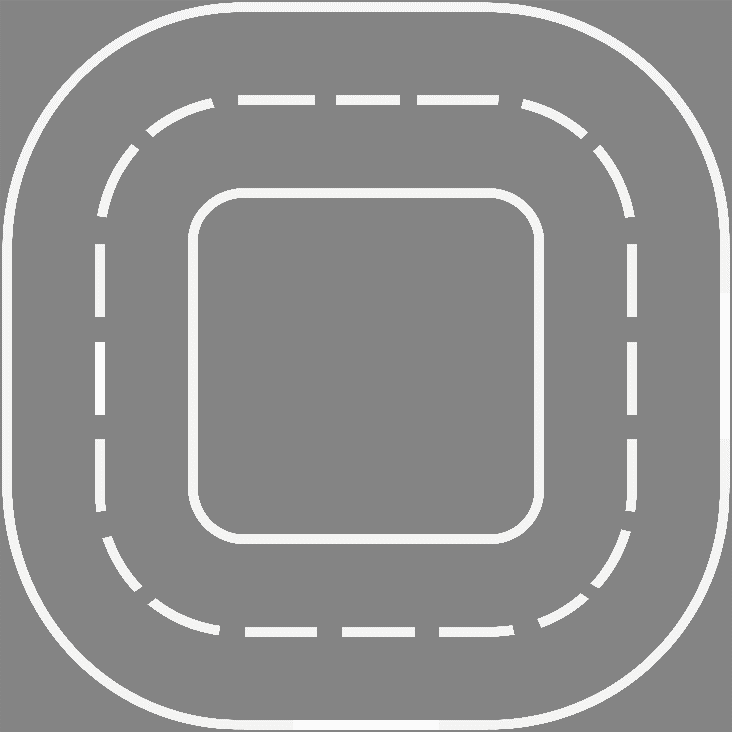}
    \caption{Experimental map on which the robot vehicle will be
moved.}
    \label{fig:fig3}
\end{figure}

In the Supervised Learning approach, the data should be
given to the trainer as x and y outputs. While the robot
vehicle was being moved over the experimental
environment, the servo angle information, which is the
current action information, was recorded in the filing system
along with the camera image. According to the general
structure of the experimental map, the servo x angle values
in the data collected vary between 1270 and 600.

\begin{figure}[h]
    \centering
    \includegraphics[scale=0.27]{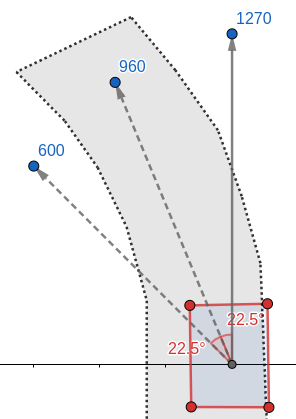}
    \caption{The angles the robot vehicle takes at the moment of
movement.}
    \label{fig:fig4}
\end{figure}

Angle data collected at this stage has a complex structure and
needs to be simplified. For this purpose, while the robot
vehicle is moving on the experimental map, the angle
information as FLAT, MIDDLE, and SHARP is updated in
the filing system by simultaneously looking at its location on
the map and the instant angle information from the computer.
Alternatively, the angle information, which is a parameter of
the data set, can be compressed between 0 and 1 for linear
regression \cite{edwards1984introduction}, \cite{montgomery2021introduction}, 
allowing linear estimation.

\subsubsection{Data Set Collection and Editing}

While the robot vehicle is in motion, it should analyze the
environmental conditions and make control predictions.
Environmental conditions consist of data collected in the
previous topic. The robot vehicle needs a system that can use
this data and make predictions. Artificial neural networks are Artificial Intelligence
structures that are trained with the given data and can make
predictions according to the information they learn.

In this subsection, the design of the architecture and the
training of the model presented initially. Next, the hyper
parameter tuning \cite{bergstra2011algorithms}, \cite{feurer2019hyperparameter}, 
and data augmentation \cite{mikolajczyk2018data}, and
finally the testing and result are explained in detail.

\subsubsection{Designing and Training the Model}

Before creating a neural network model, the neural network
structure to be used is decided by considering the data set,
project conditions and properties. The main source of data
consists of images. The neural network is required to be
predicted according to the images and action information
taken from the camera. The neural network will distinguish
the features in the images taken from the camera and perform
the learning and prediction processes. Therefore, it was
decided to use convolutional neural networks at this stage of
the project.

Data set consists of binary color pictures simplified with
Canny edge detection algorithm and angle information,
which is the action information at the time the picture is
taken. The stored images were resized to 128 x 128 pixels
before being transferred to the model. The action
information, FLAT, MIDDLE and SHARP, are updated to
correspond to 0, 1, and 2, respectively. Due to the general
structure of neural networks, the complexity of the structure
is directly proportional to the estimation time. Therefore, it
is important that the model to be designed has a simple
structure. On the other hand, the education period of the
models with a simple structure is short and time saving is
obtained. Another issue in neural networks is that there are
no rules for establishing the best model. For this reason, until
we find the model with which we have achieved high
performance, the models have been designed by taking the
available data into consideration.

The steps to be taken during the training of our artificial
neural network model are as follows. The first step is to read
and store the data set and mix it randomly. The second step
is to separate 70
test data. After separating the training and test data set, an
image from the training data set is given to our model and
the weights are updated according to the error value. Figure
5. shows the flow chart representing the training process of
the model.

\begin{figure}[h]
    \centering
    \includegraphics[scale=0.35]{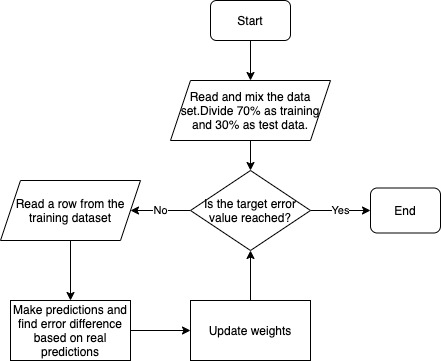}
    \caption{The algorithm flow chart representing the training
process of neural network model.}
    \label{fig:fig5}
\end{figure}

The first layer of the first model prepared is a convolution
layer with 32 x 32 depth and 3 x 3 filter dimensions. Input
data is 128 x 128 x 2 size simplified image with Canny edge
detection algorithm. Relu, the next layer activation function,
has been applied. Two Max-Pooling layers were then
applied. Filter dimensions of the Max-Pooling layers are
determined as 2 x 2. By applying Max-Pooling layers in
succession, which yields a feature map of the 32 × 32 × 32
size. Next, the Flatten layer added and then the Dropout layer
have been added to prevent overfitting \cite{srivastava2014dropout}.

\begin{figure}[h]
    \centering
    \includegraphics[scale=0.25]{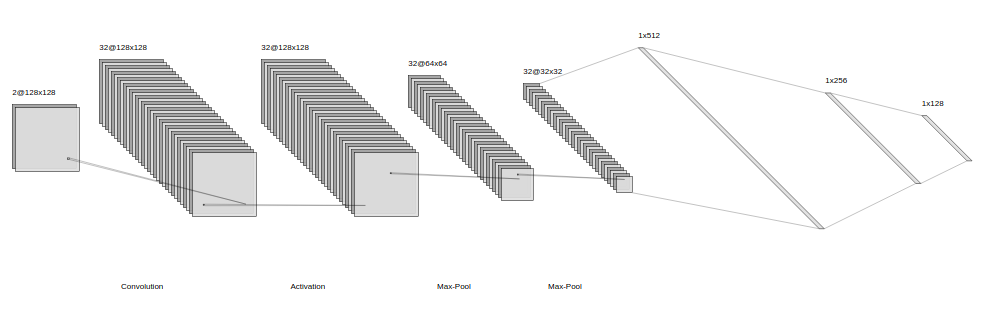}
    \caption{First CNN model architecture.}
    \label{fig:fig6}
\end{figure}

The dropout layer is a regularization approach \cite{ciliberto2016consistent} that helps
reduce dependent learning between neurons. Figure 7. shows
the loss graph of the first designed neural network.

\begin{figure}[h]
    \centering
    \includegraphics[scale=0.65]{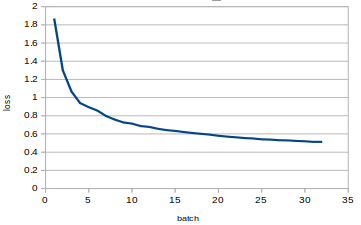}
    \caption{Loss graph of the first CNN model.}
    \label{fig:fig7}
\end{figure}

When the loss graph in Figure 7. is examined, it is seen that
during the batch of 32 pictures each, it goes to overfitting
quickly \cite{hawkins2004problem}. It has been observed that the loss of the neural
network model rapidly approaches 0 at the end of 1 epoch.

The result we will get here is that the dropout layer used to
reduce overfitting is insufficient. In order to eliminate this
problem caused by the fact that the dataset consists of few
and similar images, data diversity will be increased by data
augmentation. Thus, a more general model that can respond
to real-life problems will be obtained by considering
parameters such as lighting conditions and noise in the
image.

\subsubsection{Hyper Parameter Tuning and Data Augmentation}

While designing a model in artificial neural networks, there
is no rule to reach a successful model. There is no rule to be
followed in line with the information obtained from the
studies conducted on this subject in the world so far. The
improvement of the model is done by techniques such as
trial-and-error method, hyper parameter tuning \cite{bergstra2011algorithms} and data
augmentation \cite{mikolajczyk2018data}.

In another neural network model prepared, the first layer is a
convolution layer with 32 x 32 depth and 3 x 3 filter
dimensions. The second layer is also a convolution layer of
16 x 16 depth and 3 x 3 filter dimensions. Relu activation
function has been applied in the convolution layers. Next
comes Flatten layer and then Dense layer, which uses the
Softmax activation function.

\begin{figure}[h]
    \centering
    \includegraphics[scale=0.5]{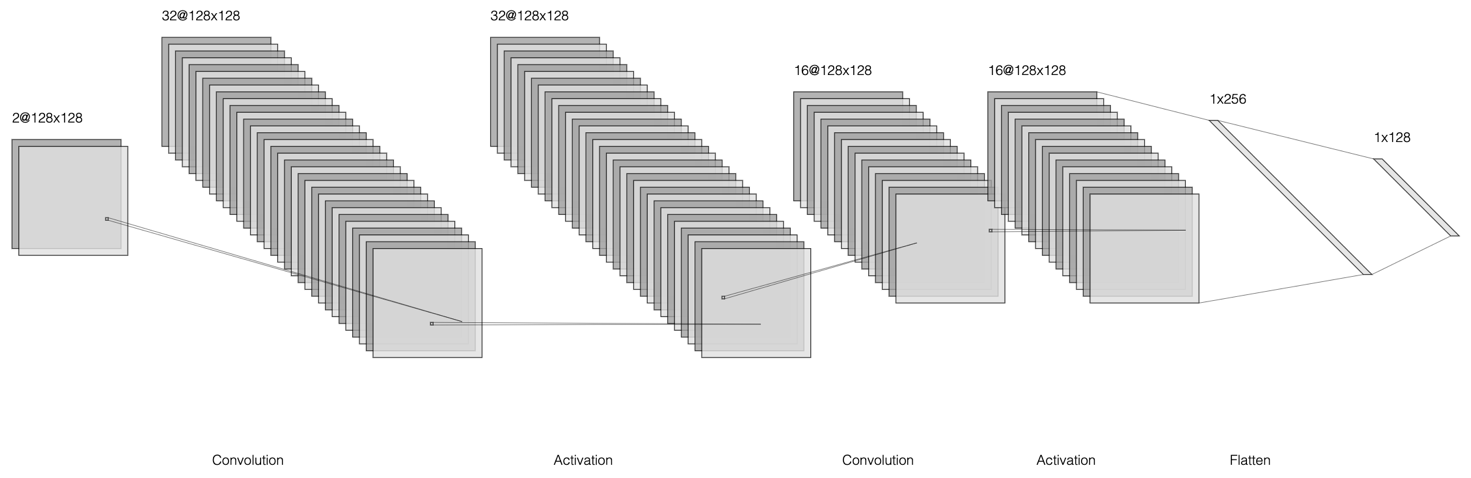}
    \caption{Layers of neural network whose hyper parameters are
tuned.}
    \label{fig:fig8}
\end{figure}

The Dropout layer was insufficient to prevent overfitting in
the previous architecture. For this reason, overfitting is
prevented by generating synthetic data with the data
augmentation technique \cite{takahashi2019data}.

\begin{figure}[h]
    \centering
    \includegraphics[scale=0.45]{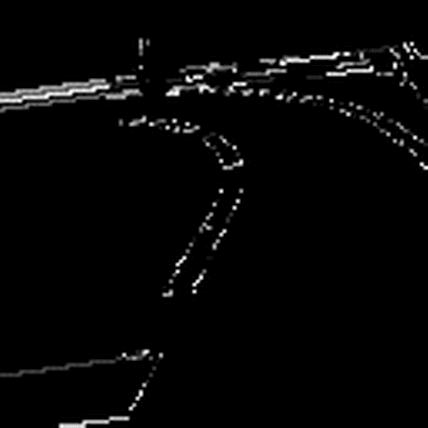}
    \caption{Synthetic image created through data augmentation.}
    \label{fig:fig9}
\end{figure}

Figure 9. shows the synthetic image obtained after applying
flip, shift, and zoom to the real image. These variations in
the data set enable the trained model to achieve a similar
performance in different image conditions. Thus, a more
general solution will be achieved. The loss graph of the
model trained after data augmentation is shown in Figure 10.

\begin{figure}[h]
    \centering
    \includegraphics[scale=0.5]{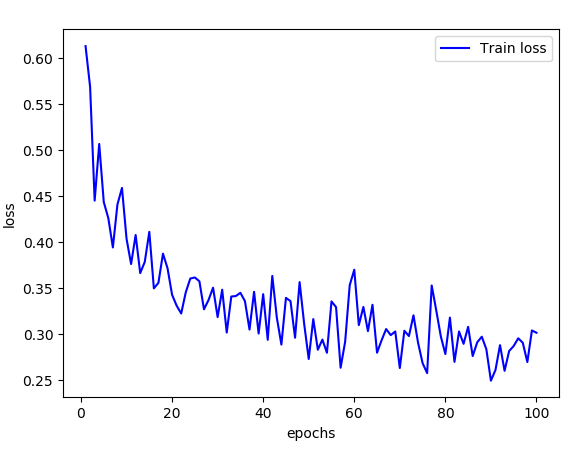}
    \caption{Loss graph of the model trained after data
augmentation.}
    \label{fig:fig10}
\end{figure}

There is a obvious improvement in overfitting \cite{hawkins2004problem} rate
compared to previous training. It is seen that the loss ratio
that converges rapidly to 0 at the batch level before is now
decreasing at the epoch level. At the end of 100 epoch, the
lowest loss value was reached as 0.25 in the 90th epoch.

New synthetic data were generated by making changes in the
augmentation parameters to reduce the loss value even
smaller. Zoom ratio decreased from 0.4 to 0.2, flip angle
from 40 degrees to 10 degrees. The changes applied here will
make less distortion of the lane information in the image and
help achieve the goal of obtaining a more general model. The
highest performing neural network model has been retrained
with the new dataset, and the expected reduction in loss data
occurred.

\begin{figure}[h]
    \centering
    \includegraphics[scale=0.5]{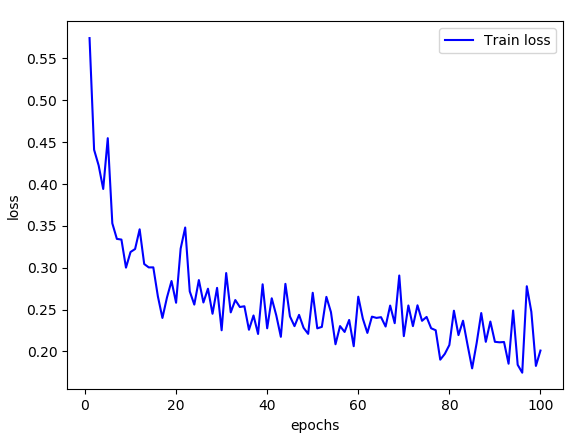}
    \caption{Loss graph of the most successful CNN model.}
    \label{fig:fig11}
\end{figure}

As seen in the graph, the 95th epoch has reached 0.17 loss
value. Thus, it was observed that the change made in the data
augmentation parameters had an effect on the decrease of the
loss value.

\subsection{Obstacle Avoidance with Optical Flow}

While an autonomous robot vehicle is moving in a constant
velocity, the time until the collision can be found without any
knowledge of the distance to be traveled or the velocity the
robot is moving \cite{o2005optical}. Calculating the time to collision is one
of the practical optical flow uses. The optical flow
knowledge is extracted from the image sequence taken from
the Google camera placed in the robotic vehicle, and then the
time until the robot reaches a particular area is determined.
Calculated collision times are considered separately as
collision times on the left and right of the image. Depending
on whether the difference between the collision times of the
left and right side is higher or lower than a certain threshold
value, the vehicle is ordered to ignore the obstacle in front of
it or to take action.

\break

\begin{figure}[h]
    \centering
    \includegraphics[scale=0.5]{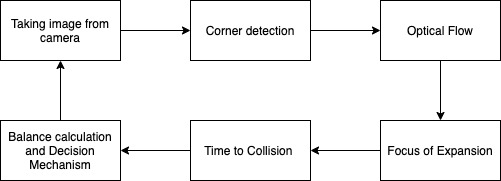}
    \caption{The flow char representing obstacle avoidance
procedure \cite{souhila2007optical}.}
    \label{fig:fig12}
\end{figure}

In this section, corner detection, and calculating optical flow
are introduced in the first place. After, the focus of expansion
and time to collision calculation procedures are explained.
Next, the balance strategy and decision mechanism are
explained in detail and the movement of the vehicle is
presented according to the decision produced by the
mechanism.

\subsubsection{Corner Detection with Fast (Features from Accelerated Segment Test)}

t is necessary to extract the optical flow information from
the image sequence taken from the camera. To find the
optical flow between consecutive frames, the motion of a
pixel feature set should be tracked. Features in the image are
points of interest that provide rich picture content
information, and these points are not affected by intensity
changes in the image \cite{fleet2006optical}.

Using the FAST \cite{viswanathan2009features} algorithm, which is known for its high
performance in real-time images, corner detection performed
in the real-time image sequence.

\begin{figure}[h]
    \centering
    \includegraphics[scale=0.33]{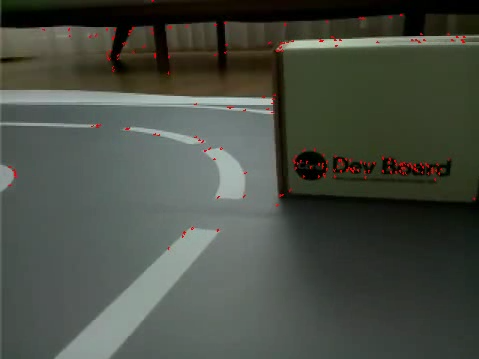}
    \caption{Corner detection on the image with FAST algorithm.}
    \label{fig:fig13}
\end{figure}

Figure 13. shows the image formed after applying the corner
detection algorithm on the image. The corner detection
process is run every 50th iteration of the runtime, which
means that the corners are refreshed at approximately 3-4
second intervals. The detected corners are stored on a vector
for later use in calculating the optical flow.

\subsubsection{Calculating Optical Flow}

For the optical flow to be computable, a selected point on the
first image must change its location on the next image. While
the selected point is moving, the shape of the light reflected
on that point is constantly changing and optical flow occurs.
In other words, the vehicle must be moving in order to obtain
optical flow with the robot vehicle.

The most widely used Lucas-Kanade \cite{lucas1981iterative} method was used
to calculate the optical flow between consecutive frames.
The vector containing the vertices detected by the FAST
corner detection algorithm is given to the function and it
returns two vectors containing the (x, y) coordinates of the
previous and next points. Now that the changing coordinates
of a corner point in the previous and ongoing frame are
known, an arrow can be drawn from the previous position to
the next position. In other words, an arrow is drawn in the
direction of the point's movement in consecutive frames if
the tracked corner point exists (detected) in the next frame.
Suppose ($x_1$, $y_1$ ) and ($x_2$ , $y_2$ ) are the coordinates of the point
in the previous and next squares.

\begin{equation}
	angle = arctan\left( \frac{x_2-x_1}{y_2-y_1} \right)
\end{equation}

\begin{equation}
	\begin{split}
		arrow_x=x_2+len*cos\left(angle+\frac{3.14}{180} \right)\\
		arrow_y=y_2+len*sin\left(angle+\frac{3.14}{180} \right)
	\end{split}
\end{equation}

\begin{figure}[h]
    \centering
    \includegraphics[scale=0.33]{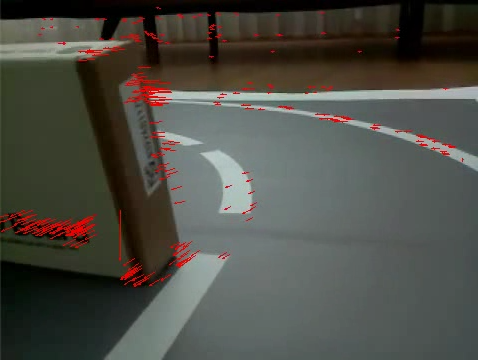}
    \caption{Arrows were drawn in the direction of movement of
the points.}
    \label{fig:fig14}
\end{figure}

\subsubsection{Calculating Focus of Expansion}

The motions of objects moving around are projected to the
eyes of the observer as two fundamental motions. An optical
flow field is formed as a result of the projection of the
translation and rotation fundamental motions into an image
plane \cite{o2005optical}. Rotational motion can be imagine as flow vectors
produced as a result of the surrounding objects shifting left
or right as the robot vehicle turns left or right.

Translation motion occurs when the camera is moving
forward or backward. If the camera moves backward, it
creates an area called a focus of contraction (FOC) where the
flow vectors converging around a point. On the contrary, if
it moves forward, it creates an area called the focus of
expansion (FOE) where the flow vectors diverge around
from a central point.

\begin{figure}[h]
    \centering
    \includegraphics[scale=0.25]{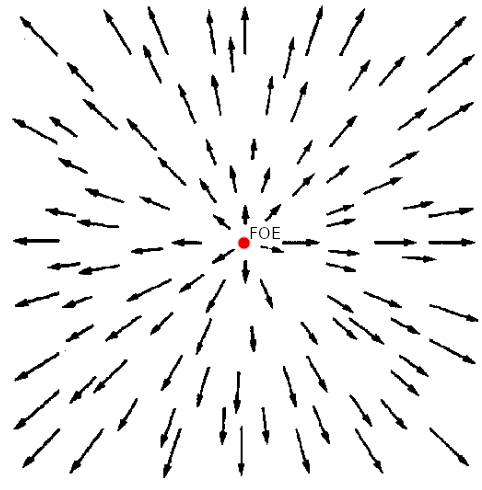}
    \caption{Diverging flow vectors and focus of expansion
during forward translation motion.}
    \label{fig:fig15}
\end{figure}

Any two vectors are needed to calculate the focus of
expansion. If the place where these two vectors meet can be
determined, the focus of expansion is found. The least-
squares \cite{levenberg1944method} solution of all available flow vectors was used
to find focus of expansion. Each optical flow vector has a
previous point and delta. Let pt = (x, y) be the x and y
coordinates of the previous position of an optical flow
vector. Let v = (u, v) be the x and y coordinate differences
between the previous and current position of the optical flow
vector.

\begin{equation}
	A=\begin{bmatrix}
		a_{00} & a_{01}\\
		...  & ... \\
		a_{n0} & a_{01}
	\end{bmatrix} \quad
	b=\begin{bmatrix}
		b_0\\
		..  \\
		b_0
	\end{bmatrix}
\end{equation}

In matrix A it should be known as a $i_0$ = -v and a $i_1$ = u, and
in matrix B, each value is obtained by $b_i$ = xv – yu. The focus
of expansion is calculated using the least-squares method
and inversion of the matrices.

\begin{equation}
	FOE=(A^TA)^{-1}A^Tb
\end{equation}

\begin{equation}
\begin{split}
=\begin{bmatrix}
		\sum _{i=1}^{N} \alpha _{i0}\beta _{i}\sum _{j=1}^{N} \alpha _{j1}^2 - \sum _{i=1}^{N} \alpha _{i1}\beta _{i}\sum _{j=1}^{N} \alpha _{j0}\alpha _{j1}\\
		- \sum _{i=1}^{N} \alpha _{i0}\beta _{i}\sum _{j=1}^{N} \alpha _{j0}\alpha _{j1} + \sum _{i=1}^{N} \alpha _{i1}\beta _{i}\sum _{j=1}^{N} \alpha _{j0}^2 \end{bmatrix}\\
- \frac{1}{\sum _{j=1}^{N} \alpha _{j0}^2\alpha _{j1}^2 - (\sum _{i=1}^{N} \alpha _{i0}\alpha _{i1})^2}
\end{split}
\end{equation}

The OpenCV library has the necessary functionality for the
matrix inversion method. The function is given matrices A
and b and an empty matrix of 2x1 dimensions. Additionally,
the DECOMP\_QR flag was added for QR decomposition \cite{gander1980algorithms}.

\begin{figure}[h]
    \centering
    \includegraphics[scale=0.33]{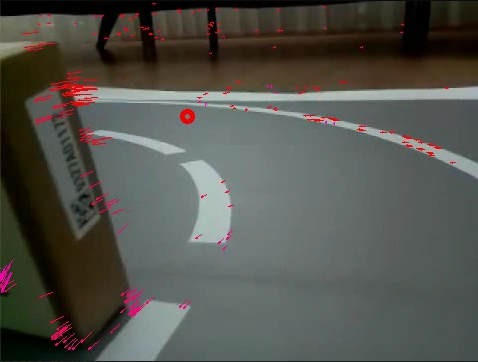}
    \caption{Calculation of the focus of expansion on the image
taken from the camera. The FOE is shown by the red circle
in the image.}
    \label{fig:fig16}
\end{figure}

\subsubsection{Calculating Time to Collision}

The most valuable information we can obtain for the robot
vehicle to avoid obstacles is the determination of the contact
time or the time until the collision. This information can be
found without requiring any information about the distance
to travel or the velocity the robot is moving [8].

The studies carried out up to this stage were to obtain
information to be used in the TTC because when calculating
the TTC, optical flow vectors and FOE are required. Let p =
(x, y) be the x and y positions of an optical flow vector and
FOE = (x, y) be the x and y positions of a focus of expansion.
Let v = (u, v) be the x and y coordinate differences between
the previous and current positions of the optical flow vector.

\begin{equation}
	TTC = \sqrt{\frac{(p_x-foe_x)^2+(p_y-foe_y)^2}{u^2+v^2}}
\end{equation}

The locations of the detected points on the image vary
according to the position and movement of the objects
captured by the robot's camera. There are situations where
the detected points on the image are not equably positioned
in the whole image plane. This means there are more corners
in some parts of the image and fewer corners in others. This
imbalance caused by the objects and movements in the frame
can be avoided by using 16 x 16 dimensional matrices.

\begin{figure}[h]
    \centering
    \includegraphics[scale=0.34]{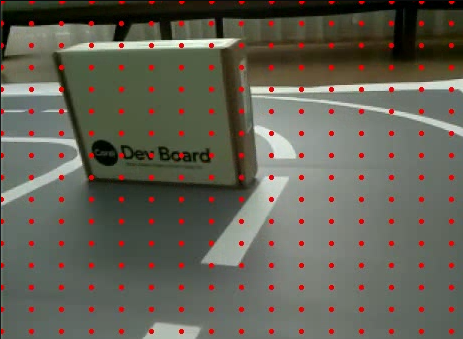}
    \caption{Drawing the matrix on the image.}
    \label{fig:fig17}
\end{figure}

After calculating the TTC of each flow vector, it is collected
at one of the closest A matrix points in the image. In matrix
B, the number of TTCs collected at each matrix point is
stored.

\begin{equation}
	A=\begin{bmatrix}
		\sum ttc_i & \dots & \sum ttc_i\\
		.  & . & . \\
		.  & . & . \\
		\sum ttc_i & \dots & \sum ttc_i
	\end{bmatrix}_{16x16} \quad
	b=\begin{bmatrix}
		\alpha _{00} & \dots & \alpha _{0n}\\
		.  & . & . \\
		.  & . & . \\
		\alpha _{n0}  & \dots & \alpha _{nn} 
	\end{bmatrix}_{16x16}
\end{equation}

ttc i is the time until the collision of a flow vector, and a in
matrix B is the number of flow vectors. Using matrices A
and b, the average TTC for each matrix point can be
calculated.

\begin{equation}
	TTC[i] = \frac{A[i]}{b[i]}
\end{equation}

\begin{figure}[h]
    \centering
    \includegraphics[scale=0.34]{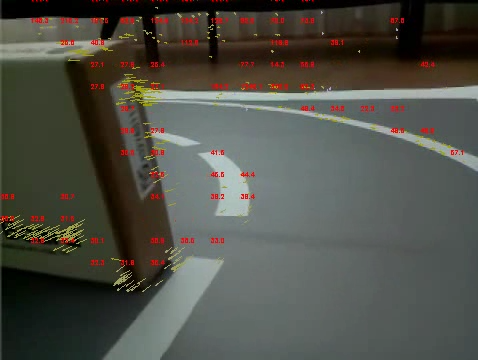}
    \caption{Shows the average collision times. It seen that
vectors with the large optical flow on the image have low
TTC values.}
    \label{fig:fig18}
\end{figure}

\subsubsection{Balance Calculation and Decision Mechanism}

The basic idea is that when the robot is in motion, close
objects move faster than farther objects on the retina. Also,
closer objects cover the field of view more, causing greater
optical flows. In the region where the optical flow is greater,
the collision time is low and the robot vehicle must go to the
other side and avoid from the obstacle. Kachluche Souhila
and Achour Karim in their article [22], they present a
different perspective in which the robot car moves away
from the side where there is greater optical flow.

The image taken from the camera is divided into two parts to
give the vehicle balance. Balance can be achieved by
minimizing the difference between collision times on the left
and right side of the vehicle.

\begin{figure}[h]
    \centering
    \includegraphics[scale=0.34]{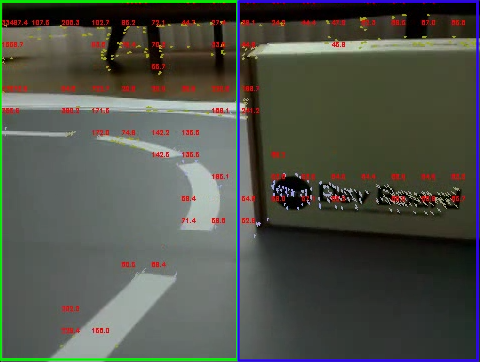}
    \caption{Dividing the image into two parts.}
    \label{fig:fig19}
\end{figure}

The following control formula is used to calculate the
difference between collision times on the left and right side
of the vehicle.

\begin{equation}
	\Delta(F_L-F_R) = \frac{\sum |TTC_L| - \sum |TTC_R|}{\sum |TTC_L| + \sum |TTC_R|}
\end{equation}

Here ($F_L$ - $F_R$ ) is the difference between the forces on both
sides of the robot body and TTC is the average of the
collision time in the visual half-field on one side. The
difference between the forces calculated in equation 8 is a
linear number and varies between 0 and 1.The balance
mechanism was applied to the robot vehicle. Due to the
environmental conditions of the experimental environment,
the threshold value of the difference between the left and
right TTC was determined as 0.5. In cases where the
threshold value is exceeded, the robot vehicle will be given
the necessary rotation order. As shown in Figure 21, the left
collision time is calculated as (3895.2), the right collision
time (101820.7) and the difference in forces (0.9) were
found. It is decided to turn right, because the left collision
time is less than the right, and the difference in forces is
greater than 0.5.

\begin{figure}[h]
    \centering
    \includegraphics[scale=0.34]{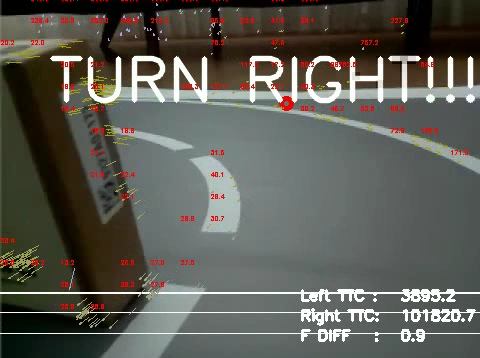}
    \caption{Avoiding the box to the left of the robot vehicle.}
    \label{fig:fig20}
\end{figure}

In Figure 21, it can be seen that the robot vehicle is given a
forward motion command. Left collision time (993.2), right
collision time (339.8) and difference in forces (0.1) were
found.

\begin{figure}[h]
    \centering
    \includegraphics[scale=0.34]{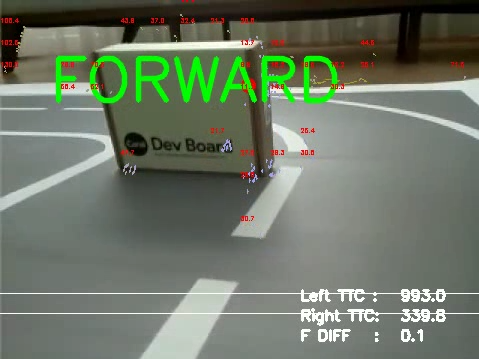}
    \caption{Robot vehicle is commanded to go forward.}
    \label{fig:fig21}
\end{figure}

\section{Conclusion}

This article presented develop lane tracking and obstacle
avoidance capabilities with different methods and solutions
for an autonomous driving system on an experimental
vehicle.

Training of the neural network model was done with
simplified images. During the test stage, the images taken
from the camera should be similar to the images used in the
training stage of the neural network. The similarity
emphasized here is that the images are in the same color
space and simplified. For this reason, the images taken from
the camera during the test stage are instantly simplified and
then transmitted to the neural network. The prepared neural
network model produces 0, 1, and 2 as output. These
correspond to the values for FLAT, MEDIUM, and SHARP,
respectively. Since these labels are obtained by simplifying
the rotation angle of the servo, the rotational motion is
provided by converting the predictions back to the servo
angle with the help of an algorithm.

Neural network models were tested on the experimental map
and the output analyzed. Although the first model did not
apply data augmentation and was less trained than other
models, it achieved 50
trained after hyper parameter adjustment and data
augmentation was observed to complete the map completely.
Considering the light conditions on the map and the external
environmental factors detected by the camera, it has been
observed that the data augmentation process eliminates this
problem and helps the vehicle to complete the map under
different light and environmental conditions.

Time to Collision, which is the target problem of obstacle
avoidance, was calculated. For this purpose, corner
detection, optical flow, and focus of expansion were
calculated, respectively. Finally, balance calculation were
made for the left and right body of the robot vehicle and a
decision mechanism was created to avoid obstacles.


\footnotesize
\section*{Acknowledgements}
This work was supported by the
Scientific and Technological Research Council of Turkey
(TÜBİTAK), Grant No: 1919B011903963

\normalsize


\end{document}